# GREY LEVEL CO-OCCURRENCE MATRICES: GENERALISATION AND SOME NEW FEATURES

Bino Sebastian V[1], A. Unnikrishnan[2] and Kannan Balakrishnan[1]

[1]Department of Computer Applications, Cochin University of Science and Technology, Cochin
binosebastianv@gmail.com, mullayilkannan@gmail.com

[2]Scientist 'G', Associate Director, Naval Physical and Oceanographic Laboratory, Cochin
unnikrishnan_a@live.com

## ABSTRACT

*Grey Level Co-occurrence Matrices (GLCM) are one of the earliest techniques used for image texture analysis. In this paper we defined a new feature called trace extracted from the GLCM and its implications in texture analysis are discussed in the context of Content Based Image Retrieval (CBIR). The theoretical extension of GLCM to n-dimensional gray scale images are also discussed. The results indicate that trace features outperform Haralick features when applied to CBIR.*

## 1. INTRODUCTION

Texture is an important characteristics used in identifying regions of interest in an image. Grey Level Co-occurrence Matrices (GLCM) is one of the earliest methods for texture feature extraction proposed by Haralick et.al. [1] back in 1973. Since then it has been widely used in many texture analysis applications and remained to be an important feature extraction method in the domain of texture analysis. Fourteen features were extracted by Haralick from the GLCMs to characterize texture [2]. Many quantitative measures of texture are found in the literature [3, 4, 5,6]. Dacheng et.al.[7] used 3D co-occurrence matrices in CBIR applications. Kovalev and Petrov [8] used special multidimensional co-occurrence matrices for object recognition and matching. Multi dimensional texture analysis was introduced in [9], which is used in clustering techniques. The objective of this work is to generalize the concept of co-occurrence matrices to n-dimensional Euclidean spaces and to extract more features from the matrix. The newly defined features are found to be useful in CBIR applications. This paper is organized as follows. The theoretical development is presented in section 2, where the generalized co-occurrence matrices and trace are defined and the numbers of possible co-occurrence matrices are evaluated. Section 3 illustrates the use of trace in CBIR by comparing its performance with the Haralick features. Section 4 concludes the paper illustrating the future works.

## 2. THEORETICAL BACKGROUND

In 1973 Haralick introduced the co-occurrence matrix and texture features for automated classification of rocks into six categories [1 ]. These features are widely used for different kinds of images. Now we will explore the definitions and background needed to understand the computation of GLCM.

### 2.1. Construction of the Traditional Co-occurrence Matrices

Let I be a given grey scale image. Let N be the total number of grey levels in the image. The Grey Level Co-occurrence Matrix defined by Haralick is a square matrix G of order N, where the $(i, j)^{th}$ entry of G represents the number of occasions a pixel with intensity i is adjacent to a pixel with intensity j. The normalized co-occurrence matrix is obtained by dividing each element of G by the total number of co-occurrence pairs in G. The adjacency can be defined to take place in each of the four directions (horizontal, vertical, left and right diagonal) as shown in figure1. The Haralick texture features are calculated for each of these directions of adjacency [10].

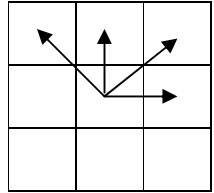

Figure 1. The four directions of adjacency for calculating the Haralick texture features

The texture features are calculated by averaging over the four directional co-occurrence matrices. To extend these concepts to n-dimensional Euclidean space, we precisely define grey scale images in n-dimensional space and the above mentioned directions of adjacency in n-dimensional images.

### 2.2. Generalized Gray Scale Images

In order to extend the concept of co-occurrence matrices to n-dimensional Euclidean space, a mathematical model for the above concepts is required. We treat our universal set as $Z^n$. Here $Z^n = Z \times Z \times \ldots \times Z$, the Cartesian product of Z taken n times with itself. Where, Z is the set of all integers. A point (or pixel in $Z^n$) X in $Z^n$ is an n-tuple of the form X=$(x_1, x_2, \ldots, x_n)$ where $x_i \in Z$ $\forall i = 1, 2, 3 \ldots n$. An image I is a function from a subset of $Z^n$ to Z. That is $f : I \to Z$ where $I \subset Z^n$. If $X \in I$, then X is assigned an integer Y such that $Y = f(X)$. Y is called the intensity of the pixel X. The image is called a grey scale image in the n-dimensional space $Z^n$. Volumetric data [11] can be treated as three dimensional images or images in $Z^3$.

### 2.3. Generalized Co-occurrence Matrices

Consider a grey scale image I defined in $Z^n$. The gray level co-occurrence matrix is defined to be a square matrix $G_d$ of size N where, N is the N be the total number of grey levels in the image. the $(i, j)^{th}$ entry of $G_d$ represents the number of times a pixel X with intensity value i is separated from a pixel Y with intensity value j at a particular distance k in a particular direction d. where the distance k is a nonnegative integer and the direction d is specified by $d = (d_1, d_2, d_3, ..., d_n)$, where $d_i \in \{0, k, -k\} \ \forall i = 1, 2, 3, ..., n$.

As an illustration consider the grey scale image in $Z^3$ with the four intensity values 0, 1, 2 and 3. The image is represented as a three dimensional matrix of size $3 \times 3 \times 3$ in which the three slices are as follows.

$$\begin{bmatrix} 0 & 0 & 1 \\ 0 & 1 & 2 \\ 0 & 2 & 3 \end{bmatrix}, \begin{bmatrix} 1 & 2 & 3 \\ 0 & 2 & 3 \\ 0 & 1 & 2 \end{bmatrix} \text{ and } \begin{bmatrix} 1 & 3 & 0 \\ 0 & 3 & 1 \\ 3 & 2 & 1 \end{bmatrix}$$

The three dimensional co-occurrence matrix $G_d$ for this image in the direction $d = (1, 0, 0)$ is the $4 \times 4$ matrix

$$G_d = \begin{bmatrix} 1 & 3 & 2 & 1 \\ 0 & 0 & 3 & 1 \\ 0 & 1 & 0 & 3 \\ 1 & 1 & 1 & 0 \end{bmatrix}$$

Note that

$$G_{-d} = \begin{bmatrix} 1 & 0 & 0 & 1 \\ 3 & 0 & 1 & 1 \\ 2 & 3 & 0 & 1 \\ 1 & 1 & 3 & 0 \end{bmatrix} = G_d{'}$$

It can be seen that $X + d = Y$, so that $G_{-d} = G_d{'}$, where $G_d{'}$ is the transpose of $G_d$. Hence $G_d + G_{-d}$ is a symmetric matrix. Since $G_{-d} = G_d{'}$, we say that $G_d$ and $G_{-d}$ are dependent (or not independent). Therefore the directions d and –d are called dependent or not independent.

Theorem: If $X \in Z^n$, the number of independent directions from X in $Z^n$ is $\frac{3^n - 1}{2}$.

Proof: Suppose $X \in Z^n$. If $Y \in Z^n$ is such that X+d=Y, where $d = (d_1, d_2, d_3, ..., d_n)$. We know that $d_i \in \{0, k, -k\}$, if the distance between X and Y is k. So we need to count the number of possibilities for forming the direction d. There are n positions $d_1, d_2, d_3, ..., d_n$ each of which can be filled using any of the three numbers 0, k or –k. This can be done in $3^n$ ways by multiplication principle. When all the positions are filled using 0, we have $d = (0, 0, 0, ..., 0)$ so that X+d=Y implies X=Y. Therefore there are $3^n - 1$ directions from X in which exactly half of the directions are independent. Therefore there are $\frac{3^n - 1}{2}$ independent directions from X in $Z^n$.

If two directions are independent, the corresponding co-occurrence matrices are transposes of each other. The above theorem indicates that the number of possible co-occurrence matrices for an n-dimensional image is $\frac{3^n - 1}{2}$.

## 2.4. Normalized Co-Occurrence Matrix

Consider $N = \sum_i \sum_j G_d(i,j)$, which is the total number of co-occurrence pairs in $G_d$. Let $GN_d(i,j) = \frac{1}{N} G_d(i,j)$. $GN_d$ is called the normalized co-occurrence matrix, where the (i, j)th entry of $GN_d(i,j)$ is the joint probability of co-occurrences of pixels with intensity i and pixels with intensity j separated by a distance k, in a particular direction d.

## 2.4. Trace

In addition to the well known Haralick features such as Angular Second Moment, Contrast, Correlation etc. listed in [1], we define a new feature from the normalized co-occurrence matrix, which can be used to identify constant regions in an image. For convenience we consider n=2, so that the image is a two dimensional grey scale image and the normalized co-occurrence matrix becomes the traditional Grey Level Co-occurrence Matrix.

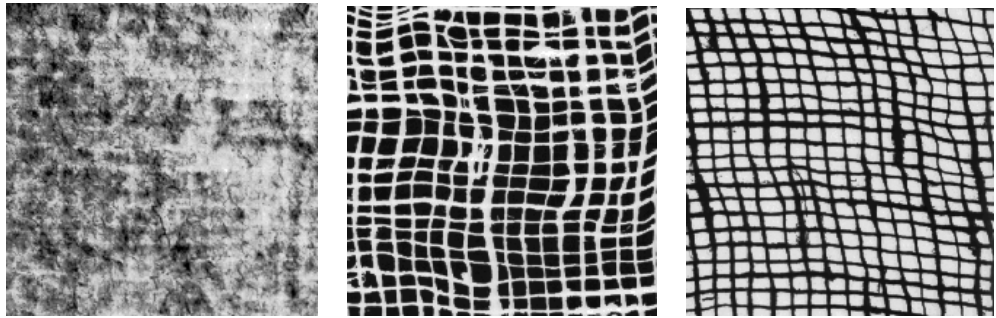

Figure 2. Sample images taken from Brodatz texture album

Consider the images taken from the Brodatz texture album given in figure 2. The majority of the nonzero entries of the co-occurrence matrices lie along the main diagonal [12] so that we treat the trace (sum of the main diagonal entries) of the normalized co-occurrence matrix as a new feature. Trace of $GN_d(i,j)$ is defined as

$$Trace = \sum_i GN_d(i,i)$$

From the definition of the co-occurrence matrix, it can be seen that an entry in the main diagonal is the $(i, i)^{th}$ entry. This implies two pixels with the same intensity value i occur together. Thus higher values of trace implies more constant region in the image. The computed values of the trace of the normalized co-occurrence matrices in Figure 2 with k=1 are 0.0682, 0.2253 and 0.2335 for the left, middle and right images respectively. Obviously the left image contains less

154

amount of constant region and the other two images contain almost the same amount of constant regions. The value of the trace indicates the same.

## 3. METHODOLOGY

Here we present the use of trace in content based image retrieval. The goal of image retrieval is to compare a given query image with all potential target images in order to obtain numerical measures of their similarity with the query image. Our database contains 333 images taken from the Brodatz texture album which contains 36 classes, each class consisting of 9 images. Retrieval results are evaluated by calculating average precision. Precision is the proportion of retrieved images that are relevant to the query.

### 3.1. Image Retrieval Using Trace

The numerical value of trace provides only a measure of the amount of constant region in an image. Thus we divide the main diagonal entries of the co-occurrence matrix into four equal parts and the sum of the elements in each quarter is taken to be a measure of the image texture feature for image retrieval, giving a four dimensional vector. The database is queried using the first and the fourth images from all the 36 different classes. Eight images are retrieved in each run. The average precision is found to be 0.8194.

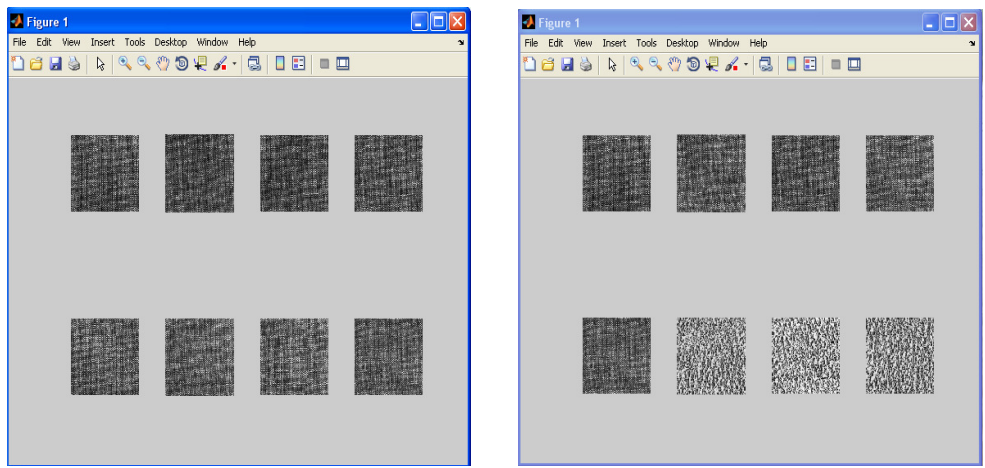

Figure3. Screen shots of the output for the same query image using the trace features (left) and the Haralick features (right)

### 3.2. Comparison of Results with Haralick features

Conducting the same experiment using the well known Haralick features Contrast, Correlation, Energy and Homogeneity we obtain an average precision of 0.7222. Here also we use a four dimensional feature vector for querying the database. This is a clear indication of the improvement of performance using the proposed features.

## 4. CONCLUSION

This paper illustrates the possible theoretical extensions of Grey Level Co-occurrence Matrices. The use of trace in texture analysis is found to be promising. Trace itself can be used as a feature which outperforms the Haralick features. Trace combined with Haralick features provides better results. Only one third of the images from Brodatz texture database are used for testing. Our future work is to investigate the performance of trace with the complete set of images in the database. Trace extracted from three dimensional images is also to be investigated. The use of the theoretical developments to n-dimensional Euclidean space need to explored.

**Authors**

**Bino Sebastian V**, born in 1975 received his M. Sc degree in Mathematics and M. Tech degree in Computer and Information Science from Cochin University of Science and Technology, Cochin, India in 1997 and 2003 respectively. He is currently working with Mar Athansius College, Kothamangalam, Kerala, India, as an Assistant Professor in the Department of Mathematics. He is pursuing for his Ph. D degree in the Department of Computer Applications, Cochin University of Science and Technology

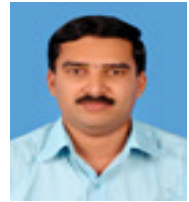

**Dr. A Unnikrishnan**, Graduated from REC (Calicut), India in Electrical Engineering(1975), completed his M. Tech from IIT, Kanpur in Electrical Engineering(1978) and PhD from IISc, Bangalore in "Image Data Structures"(1988). Presently, he is The Associate Director Naval Physical and Oceanographical Laboratory, Kochi which is a premiere Laboratory of Defence Research and Development Organisation. His field of interests include Sonar Signal Processing, Image Processing and Soft Computing. He has authored about fifty National and International Journal and Conference Papers. He is a Fellow of IETE & IEI, India.

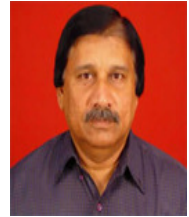

**Dr. Kannan Balakrishnan**, born in 1960, received his M. Sc and M. Phil degrees in Mathematics from University of Kerala, India, M. Tech degree in Computer and Information Science from Cochin University of Science & Technology, Cochin, India and Ph. D in Futures Studies from University of Kerala, India in 1982, 1983, 1988 and 2006 respectively. He is currently working with Cochin University of Science & Technology, Cochin, India, as an Associate Professor in the Department of Computer Applications. Also he is the co investigator of Indo-Slovenian joint research project by Department of Science and Technology, Government of India. He has published several papers in international journals and national and international conference proceedings. His present areas of interest are Graph Algorithms, Intelligent systems, Image processing, CBIR and Machine Translation. He is a reviewer of American Mathematical Reviews and several other journals.

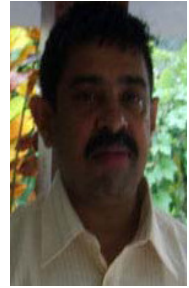